\documentclass[journal]{IEEEtran}
\usepackage{ifpdf}
\usepackage{graphicx}
\ifpdf
 \graphicspath{{Figs/}{Figs/PDF/}}
\else
 \graphicspath{{Figs/}}
\fi

\usepackage{hhline}

\usepackage{calc}
\usepackage{enumitem}

\usepackage{hyperref}
\hypersetup{
	colorlinks = true,
	linkcolor = {blue},
	citecolor = {blue}
}
\usepackage{url}

\usepackage{nomencl}
\usepackage{multirow}

\usepackage{subcaption}

\usepackage{amsmath}

\usepackage{breqn}
\usepackage{array}
\usepackage{booktabs}

\newcolumntype{P}[1]{>{\centering\arraybackslash}p{#1}}

\usepackage[flushleft]{threeparttable}

\usepackage{cite}

\ifCLASSINFOpdf
\else
\fi
\begin{document}
%
\title{iCLAP: Shape Recognition by Combining Proprioception and Touch Sensing}
%
%
%

\author{Shan~Luo, Wenxuan Mou, Kaspar Althoefer and Hongbin Liu
\thanks{Shan Luo is with the Center for Robotics Research, Department of Informatics, King\rq s
College London, London, WC2R 2LS. He is also with the Department of Computer Science at the University of Liverpool, Liverpool L69 3BX, UK.}
\thanks{Wenxuan Mou is with the Multimedia and Vision Research Group, School of Electronic Engineering and Computer Science, Queen Mary University of London, London E1 4NS, UK.}
\thanks{Kaspar Althoefer is with School of Engineering and Materials Science, Queen Mary University of London, E1 4NS, London, UK.}
\thanks{Hongbin Liu is with the Centre for Robotics Research, Department of Informatics, King\rq s College London, London, WC2R 2LS, UK.}
\thanks{Corresponding author: S. Luo (mountluoupc@gmail.com).}
}

%
%

\markboth{Autonomous Robots,~Vol.~0, No.~0, JUN 2018}%
{Shell \MakeLowercase{\textit{et al.}}: Bare Demo of IEEEtran.cls for IEEE Journals}
%



\maketitle

\begin{abstract}
For humans, both the proprioception and touch sensing are highly utilized when performing haptic perception. However, most approaches in robotics use only either proprioceptive data or touch data in haptic object recognition. In this paper, we present a novel method named Iterative Closest Labeled Point (iCLAP) to link the kinesthetic cues and tactile patterns fundamentally and also introduce its extensions to recognize object shapes. In the training phase, the iCLAP first clusters the features of tactile readings into a codebook and assigns these features with distinct label numbers. A 4D point cloud of the object is then formed by taking the label numbers of the tactile features as an additional dimension to the 3D sensor positions; hence, the two sensing modalities are merged to achieve a synthesized perception of the touched object. Furthermore, we developed and validated hybrid fusion strategies, product based and weighted sum based, to combine decisions obtained from iCLAP and single sensing modalities. Extensive experimentation demonstrates a dramatic improvement of object recognition using the proposed methods and it shows great potential to enhance robot perception ability.
\end{abstract}


%
\IEEEpeerreviewmaketitle

\section{Introduction}
%
%
%
%
\IEEEPARstart{T}{he} sense of touch plays an important role in robot perception and many tactile sensors have been developed in the last few decades \cite{zhang2002hybrid,xie2013fiber,dahiya2013directions}. In addition to obvious applications of collision detection and avoidance, tactile sensors can be applied in multiple tasks such as object recognition \cite{luo2015tactile,pezzementi2011tactile,luo2018vitac}, dexterous manipulation \cite{kappassov2015tactile,bimbo2016hand} and localization \cite{luo2015localizing,pezzementi2011object}. The haptic object recognition can be considered at two scales, i.e., local and global shapes \cite{lederman2009haptic,torres2000translation}. The local object shape, e.g., shapes that can fit into fingertips, can be recovered or recognized by a single touch, analogous to human cutaneous sense. The global shapes, e.g., contours that extend beyond the fingertip scale, usually require the contribution of both cutaneous and kinesthetic inputs. In such case, mechanoreceptors in joints are also utilized to acquire the movement of the fingers/end-effectors in space, with the assistance of local tactile features, to recognize the object identity. The kinesthetic inputs here are similar to human proprioception that refers to the awareness of positions and movements of body parts.

In this paper, an algorithm named Iterative Closest Labeled Point (iCLAP) and its extensions are presented to incorporate tactile and kinesthetic cues for haptic shape recognition. With only tactile readings, a dictionary of tactile features can be first formed. By searching for the nearest word in the dictionary, each tactile feature is assigned a label number. A four-dimensional data point is then obtained by concatenating the 3D position of the object-tactile sensor interaction point and the word label number. In this manner, 4D point clouds of objects can be obtained from training and test data. The partial 4D point cloud obtained from test set iteratively matches with all the reference point clouds in the training set and the identity of the best-fit reference model is assigned to the test object. Furthermore, weighted sum based and product based fusion strategies have also been developed for haptic shape recognition.


The contributions of this paper are: 

\begin{itemize}
   \item A novel iCLAP algorithm that incorporates tactile readings and kinesthetic cues for object shape recognition is proposed;
   \item Extensions of iCLAP algorithm based on different fusion strategies are created to enhance the recognition performance; 
   \item Extensive experiments are conducted to demonstrate the difference in the recognition performances between different approaches.
\end{itemize}

This paper extends our previous work \cite{luo2016iterative} by including extension methods of the iCLAP algorithm and introducing more thorough experiments. The remainder of this paper is organized as follows. The literature in the haptic object shape processing is reviewed in Section II. The proposed tactile-kinesthetic shape recognition system is described in Section III. A series of the fusion approaches, both product-based and weighted sum based, are then introduced in Section IV. In Section V, the experimental setup and the data acquisition are described. The experiment results are then provided and analyzed in Section VI. Finally, in Section VII the conclusions are drawn and possible applications and future research directions are presented.

\section{Literature Review}\label{literature}
Thanks to the development of tactile sensor technologies \cite{khan2015technologies,dahiya2011towards,li2013haptics}, haptic object shape processing has received increasing attention in the recent years \cite{luo2017robotic}. Some research produced point clouds by collecting contact points to constrain the geometry of the object \cite{casselli1995robustness,allen1989haptic} whereas some others relied on tactile appearances by extracting features from tactile readings \cite{schneider2009object,pezzementi2011tactile,luonovel}. In contrast, for humans, the sense of touch consists of both kinesthetic (position and movement) and cutaneous (tactile) sensing \cite{lederman2009haptic}. Therefore, the fusion of spatial information of sensor movements and features of tactile appearance could be beneficial for object recognition tasks.

The methods based on contact points often employ techniques from computer graphics to fit the obtained cloud of contact points to a geometric model and outline the object contour. These methods were widely used by early researchers due to the low resolution of tactile sensors and prevalence of single-contact force sensors \cite{allen1989haptic,charlebois1999shape,okamura2001feature}. The resultant points from tactile readings can be fit to either super-quadric surfaces \cite{allen1989haptic} or a a polyhedral model \cite{casselli1995robustness} in order to reconstruct unknown shapes. Different from the point cloud based approaches, a non-linear model-based inversion is proposed in \cite{fearing1991using} to recover surface curvatures by using a cylindrical tactile sensor. In more recent works \cite{ibrayev2005semidifferential,jia2006surface,jia2010surface}, the curvatures at curve intersection points are analyzed and thus a patch is described through polynomial fitting. Tactile array sensors have also been utilized to obtain the spatial distribution of the object in space. In \cite{pezzementi2011object} an object representation is constructed based on mosaics of tactile measurements, in which the objects are a set of raised letter shapes. Kalman filters are applied in \cite{meier2011probabilistic} to generate 3D representations of objects from contact point clouds collected by tactile array sensors, and the objects are then classified with the Iterative Closest Point (ICP) algorithm. Through utilizing these methods, arbitrary contact shapes can be retrieved, however, it can be time consuming when exploring a large object surface as excessive contacts are required to recognize the global object shape.

Another approach is to recognize the contact shapes by extracting shape features from pressure distributions within tactile images. The image descriptors from computer vision have been applied to represent the local contact patterns, e.g., image moments \cite{pezzementi2011tactile,corradi2015bayesian,drimus2014design}, SIFT based features \cite{pezzementi2011tactile,luonovel} and raw readings \cite{schneider2009object,liu2012tactile}. However, there is only a limited number of approaches available for recovering the global object shape by analyzing pressure distributions in tactile images collected at different contact locations. One popular method is to generate a codebook of tactile features and use it to classify objects \cite{schneider2009object,luo2014rotation,luonovel,madry2014st}; a particular paradigm is the bag-of-words model. In this framework, only local contact features are taken to generate a fixed length feature occurrence vector to represent the object whereas the three-dimensional distribution information is not incorporated.

For humans, haptic perception makes use of both kinesthetic and cutaneous sensing \cite{lederman2009haptic}, which can also be beneficial to robots. In \cite{mcmath1991tactile,petriu1992active}, a series of local ``tactile probe images" is assembled and concatenated together to obtain a ``global tactile image" using 2D correlation techniques with the assistance of kinesthetic data. However, in this work tactile features are not extracted whereas raw tactile readings are utilized instead, which would bring high computational cost when investigating large object surfaces. In \cite{johnsson2007neural}, the
tensor product operation is applied to code proprioceptive and cutaneous information, using Self-Organizing Maps (SOMs) and Neural Networks. In \cite{gorges2010haptic}, the tactile and kinesthetic data are integrated by decision fusion and description fusion methods. In the former, classification is done with two sensing modalities independently and recognition results are combined into one decision afterwards. In the latter, the descriptors of kinesthetic data (finger configurations/positions) and tactile features for a palpation sequence are concatenated into one vector for classification. In other words, the information of the positions where specific tactile features are collected is lost. In both methods, the tactile and kinesthetic information is not fundamentally linked. In a similar manner, in \cite{navarro2012haptic} the tactile and kinesthetic modalities are fused in a decision fusion fashion. Both tactile features and joint configurations are clustered by SOMs and classified by ANNs separately and the classification results are merged to achieve a final decision. In a more recent work \cite{spiers2016single}, the actuator positions of robot fingers and tactile sensor values form the feature space to classify object classes using random forests but there are no exploratory motions involved, with data acquired during a single and unplanned grasp. 

In our paper, a novel iCLAP algorithm is proposed to incorporate tactile feature information into the location description of the sensor-object contact in a four dimensional space. In this way, the kinesthetic cues and tactile patterns are fundamentally linked. The experiments of classifying 20 real objects show that the classification performance is improved by up to 14.76\% by using iCLAP compared to methods based on single sensing sources and a high recognition rate up to 95.52\% can be achieved when a hybrid fusion is applied.

\section{Iterative Closest Labeled Point} \label{iclap}
\begin{figure}[htbp]
	\begin{center}
		\includegraphics[width=0.8\columnwidth]{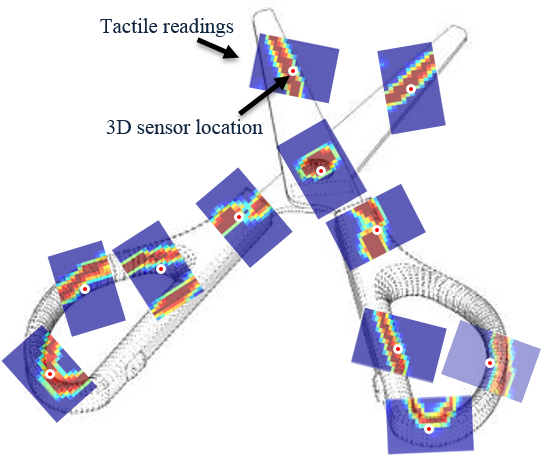}
	\end{center}
	\caption{An illustration of haptic data collection from a pair of scissors. At each data point, a tactile reading (rectangle blocks) and the 3D sensor location (red dots with white edges at the block centers) are collected.}
	\label{fig:scissor.png}
\end{figure}

In the haptic exploration, at each data collection point, both the tactile reading and the 3D sensor location can be recorded simultaneously. An illustration of data extracted from a pair of scissors is depicted in Fig.~\ref{fig:scissor.png}. Our proposed Iterative Closest Labeled Point (iCLAP) algorithm is based on two sensing sources, i.e., appearance features from obtained tactile images and spatial distributions of objects in space.

To create a label for each tactile feature, a dictionary of tactile features is first formed from training tactile readings and each tactile feature is then assigned a label by ``indexing" the dictionary. The dictionary is formed by clustering the features extracted from tactile images using $k$-means clustering, where $k$ is the dictionary size, namely, the number of clusters. The features of training/test objects are then assigned to their nearest clusters in Euclidean distance and are therefore labeled with the cluster numbers $\mu$ $\in$ \{1, 2,\ldots, $k$\}.

With the feature labels created from tactile readings and the sensor locations in 3D space, a single 4D point $p$ can be represented by a tuple $ p = (p_{x}, p_{y}, p_{z}, p_{\mu})$, where $ p_{x} $, $ p_{y} $, $ p_{z} $ are the \textit{x}, \textit{y}, \textit{z} coordinates of the tactile sensor in 3D space and $ p_{\mu} $ is the word label assigned to this location respectively. In this manner, the object can be represented in a four-dimensional space. To calculate the mutual distance between 4D sparse data point cloud \textit{P} in the test set and the model (reference) point clouds \textit{Q} in the training set, the Iterative Closest Point (ICP) algorithm \cite{besl1992method} is extended to 4D space. Let data point $ p_{i} $ and model data point $ q_{i} $ be an associated set of the \textit{N} matched point pairs. With the $4 \times 4$ rotation matrix \textit{R} and $4 \times 1$ translation vector $ \vec{t} $, $ p_{i} $ can be transformed into the coordinate system of model point cloud: $ \overline{p_{i}}=\textit{R} p_{i}+\vec{t} $. To find the closest point in the model point cloud to each transformed test data point $ \overline{p_{i}} $, a $k$-d tree of the model point cloud is constructed \cite{zhang1994iterative,bentley1975multidimensional}. An error metric $ E_{iCLAP} $ is defined to evaluate the mean square root distance of the associated point pairs and it is minimized with the optimal rigid rotation matrix \textit{R} and translation vector $ \vec{t} $. In physical terms, the error metric $ E_{iCLAP} $ can be visualized as the total potential energy of springs attached between matched point pairs. $ E_{iCLAP} $ of point clouds $ P $ and $ Q $ to be matched is $E_{iCLAP}(P,Q)=\sum_{i=1}^{N}\Vert \overline{p_{i}}-q_{i}\Vert^2$.

The centroids of \textit{P} and \textit{Q} are $c_{p}=\frac{1}{n}\sum_{i=1}^{n}p_{i} $ and $ c_{q}=\frac{1}{m}\sum_{i=1}^{m}q_{i} $, where \textit{n} and \textit{m} are the number of test and model data points respectively. In general case, \textit{n} is equal to the number of the matched point pairs \textit{N}. The point deviations from the centroids of \textit{P} and \textit{Q} can be obtained as $ {p_{i}}^{'}=p_i-c_{p} $ and $ {p_{i}}^{'}=q_i-c_{q} $. $ E_{iCLAP} $ can be rewritten as
\begin{equation}
\begin{split}
E_{iCLAP}(P,Q)&=\sum_{i=1}^{N}\Vert R({p_{i}}^{'}+c_{p})+\vec{t}-({q_{i}}^{'}+c_{q})\Vert^2\\
           &=\sum_{i=1}^{N}\Vert R{p_{i}}^{'}-{q_{i}}^{'}+(Rc_{p}-c_{q}+\vec{t})\Vert^2.
\end{split}
\end{equation}
To minimize the error metric, $ \vec{t} $ is set as $\vec{t}=c_{q}-Rc_{p}$. Therefore, the error metric is simplified as
\begin{equation}
\begin{split}
E_{iCLAP}(P,Q)&=\sum_{i=1}^{N}\Vert R{p_{i}}^{'}-{q_{i}}^{'}\Vert^2\\
           &=\sum_{i=1}^{N}\Vert {p_{i}}^{'}\Vert^2-2tr(R\sum_{i=1}^{N}{p_{i}}^{'}{{q_{i}}^{'}}^T)+\sum_{i=1}^{N}\Vert {q_{i}}^{'}\Vert^2 , 
\end{split}
\end{equation}
where \textit{R} is orthogonal for orthogonal transformation, therefore, $RR^T=\textit{I}$. Now let $ H=\sum_{i=1}^{N}{p_{i}}^{'}{{q_{i}}^{'}}^T $.
In an expanded form,
\begin{equation}
H = 
\begin{pmatrix}
M_{xx} & M_{xy} & M_{xz} & M_{x\mu} \\
M_{yx} & M_{yy} & M_{yz} & M_{y\mu} \\
M_{zx} & M_{zy} & M_{zz} & M_{z\mu} \\
M_{\mu x} & M_{\mu y} & M_{\mu z} & M_{w\mu} 
\end{pmatrix},
\end{equation}
where $ M_{ab}  $=$ \sum_{i=1}^{N} {p_{ia}}^{'}{q_{ib}}^{'}$, and $ \textit{a}, \textit{b} \in\{ x,y,z,\mu \} $. To minimize $ E_{iCLAP} $ the trace \textit{tr(RH)} has to be maximized. Let the columns of \textit{H} and the rows of \textit{R} be $ h_{j} $ and $ r_{j} $ respectively, where $ \textit{j}\in\{1,2,3,4\} $. The trace of \textit{RH} can be expanded as
\begin{equation}
tr(RH)=\sum_{j=1}^{4}r_j \cdot h_j \leq \sum_{j=1}^{4}\Vert r_j \Vert \Vert h_j \Vert ,
\end{equation}
where the inequality is just a reformulation of the Cauchy--Schwarz inequality. Since the rotation matrix \textit{R} is orthogonal, its row vectors all have unit length. This implies
\begin{equation}
tr(RH) \leq \sum_{j=1}^{4}\sqrt{h^T_j h_j}=tr(\sqrt{{H^T}H}) ,
\label{eq:maxvalue}
\end{equation}
where the square root is taken in the operator sense and $ tr(\sqrt{{H^T}H}) $ is the trace norm of $ H $ that is the sum of singular values. Consider the singular value decomposition of $ H=U\Sigma V^T $. If the rotation vector is set as $ R=VU^T$, the trace of \textit{RH} becomes
\begin{equation}
\begin{split}
tr(VU^TU\Sigma V^T)&=tr(V\Sigma V^{-1}) \\
                &=tr(\sqrt{V{\Sigma}^T\Sigma V^{-1}})\\
                &=tr(\sqrt{H^T H}) ,
\end{split}
\end{equation}
which is the maximum according to Eq.~\ref{eq:maxvalue}. It means that $ E_{iCLAP} $ is minimized with the resulting optimal rotation matrix $ R=VU^T $ and translation vector $ \vec{t}=c_{q}-Rc_{p} $.

The iCLAP is iterated until any of the termination conditions is reached: error metric $ E_{iCLAP} > $ preset tolerance; number of iterations $ > $ preset maximum number of iterations $ n_{max} $; the relative change in the error metrics of two consecutive iterations falls below a predefined threshold. The obtained distances between the test and reference point clouds are then normalized by L2 norm. A reference point cloud with the minimum $ E_{iCLAP} $ can be found and its identity is assigned to the test object by comparing $ E_{iCLAP} $.

\section{Fusion Methods for Haptic Object Recognition}
In general, the fusion of multiple modalities can provide complementary knowledge and improve the decision-making performance; it can be performed at either the feature level or the decision level \cite{atrey2010multimodal,liu2017classifier}. In feature fusion, the features extracted from different modalities are combined into high dimensional feature vectors prior to the classification, which are fed into a single classifier. In essence, our iCLAP algorithm is performed at the feature fusion level. However, due to the distinct representations of information sources, i.e., sensor locations and tactile labels in our case, normalization is hard to be performed and it needs to find the best normalization parameters by trial and error. Instead, decision fusion combines the decisions made based on individual modalities and makes a final decision in the semantic space where individual decisions usually have the same representation. Therefore, the decision level strategy is also adopted. Two different decision fusion methods are developed, i.e., weighted sum-based and product-based. To take advantage of both feature and decision fusion methods, hybrid fusion approaches are developed to combine the decisions of iCLAP and methods based on single sensing modalities. In total, nine synthesis methods, including one at feature level (iCLAP), two at decision level and six at hybrid level, are developed. There are two recognition pipelines, i.e., tactile based and kinesthetics based.

\subsubsection{Tactile based object recognition}
As described in Section~\ref{iclap}, a dictionary is formed by clustering tactile features extracted from the training tactile images using \textit{k}-means clustering. The descriptors are then assigned to their nearest clusters in Euclidean distance. In this way, both training and test objects can be represented by histograms of word occurrences $ h^{class} $ and $ h^{test} $ respectively. Therefore, the distance $ E_{BoW} $ between the test and reference objects can be computed using the histogram intersection metric of their word occurrence histograms $ h^{test } $ and $  h^{class } $ \cite{luonovel}: 
\begin{equation}
E_{BoW}(h^{test},h^{class})=1-\sum_{i=1}^{k}min(h^{test}_i,h^{class}_i),
\end{equation}
where \textit{k} is the dictionary size.

\subsubsection{Kinesthetics based object recognition}
In each exploratory procedure, the locations of contact points form a point cloud. To calculate the mutual distance between 3D sparse data point cloud $ P^{3} $ in the test set and the reference point clouds $ Q^{3} $ in the training set, the classic Iterative Closest Point (ICP) \cite{bentley1975multidimensional} algorithm is employed. Similar to the proposed iCLAP algorithm, a point-to-point error metric $ E_{ICP} $ between $ P^{3} $ and $ Q^{3} $ is defined:
\begin{equation}
\begin{split}
E_{ICP}(P^3,Q^3)&=(\sum_{i=1}^{N_p}\Vert R_{p}p_{i}+\vec{t_p})-q_{i}\Vert^2\\
&=\sum_{i=1}^{N_p}\sqrt{
\begin{aligned}
(&\overline{p_{ix}}-q_{ix})^2+(\overline{p_{iy}}-q_{iy})^2\\
& \quad \quad +(\overline{p_{iz}}-q_{iz})^2
\end{aligned}
,
}
\end{split}
\end{equation}
where $ R_{p} $ and $ \vec{t_{p}} $ denote the rotation matrix and translation vector in 3D space, respectively, and $ N_{p} $ is the number of matched point pairs. It is minimized iteratively to achieve an optimal transformation from the source points in $ P^{3} $ to corresponding target points in $ Q^{3} $. The obtained distances between the test point cloud and the reference models in the training set are then normalized by L2 norm.

\subsection{Decision fusion methods}
Two decision fusion methods are proposed to synthesize both the tactile-based and kinesthetic-based recognition results. One is to calculate the distance between the test object and reference objects by a weighted sum of the two distances obtained from the two pipelines $ ^I_BE_{+} $, which is named as the weighted sum fusion. In this method, the distances obtained from tactile sensing and sensor movements are combined in a linear fashion. It combines the scores/decisions in two modalities and ranks the combined results. Here \textit{I} and \textit{B} denote ICP and BoW respectively. Let $ w_{p1} $ be the weight assigned to the kinesthetic sensing source, $ ^I_BE_{+} $ can be formed as: 
\begin{equation}
^I_BE_{+}=w_{p1}E_{ICP}+(1-w_{p1})E_{BoW}
\label{eq:decisionfusionweight}
\end{equation}
The other is to acquire the distance by the product of $ E_{ICP} $ and $ E_{BoW} $, which is named as product fusion. This method is based on the probability analysis of the likelihoods of the object classes based on the two modalities. The distance $ _B^IE_{\times} $ can be formed as:
\begin{equation}
_B^IE_{\times}=E_{ICP} \times E_{BoW}.
\end{equation}
The identity of the reference object with the nearest distance to the test object is therefore assigned to the test object.
\subsection{Hybrid fusion methods}
To exploit the advantages of both feature and decision fusion strategies, a hybrid fusion strategy can be applied. Accordingly, the recognition results of iCLAP algorithm (feature fusion) are further integrated with decisions made by methods using single sensing modalities by a decision fusion to obtain the final decisions. In total, six hybrid fusion methods are developed, combining different decision fusion manners and various number of sensing modalities used. It can be divided into two groups with regards to the decision fusion manners. The first group is to compute the distances in a weighted sum manner of the two distances obtained from the kinesthetic only based pipeline ($ ^{I+}_IE_{+} $), tactile only based pipeline ($ ^{I+}_BE_{+} $) or both of them ($ ^{All}E_{+} $) and iCLAP algorithm. Here \textit{I+} stands for iCLAP algorithm. Let $ w_{p2} $ be the weight assigned to the kinesthetic sensing source, $ ^{I+}_IE_{+} $ can be formed as:
\begin{equation}
^{I+}_IE_{+}=w_{p2}E_{ICP}+(1-w_{p2})E_{iCLAP}.
\label{eq:hybridiclapicp}
\end{equation}
Let $ w_{w3} $ be the weight assigned to the tactile sensing source, $ ^{I+}_BE_{+} $ can be formed as:
\begin{equation}
^{I+}_BE_{+}=w_{w3}E_{BoW}+(1-w_{w3})E_{iCLAP}.
\label{eq:hybridiclapbow}
\end{equation}
Let $ w_{p4} $ and $ w_{w4} $ be the weights assigned to the kinesthetic and tactile sensing sources respectively, $ ^{All}E_{+} $ can be formed as:
\begin{equation}
^{All}E_{+}=w_{p4}E_{ICP}+w_{w4}E_{BoW}+(1-w_{p4}-w_{w4})E_{iCLAP}
\label{eq:hybridiclapicpbow}
\end{equation}
The second group is to evaluate the distance between the test object and the reference objects in a product manner of the distances obtained from the kinesthetic only based pipeline ($ ^{I+}_IE_{\times} $), tactile only based pipeline ($ ^{I+}_BE_{\times} $) or both of them ($ ^{All}E_{\times} $) and iCLAP algorithm. $ ^{I+}_IE_{\times} $, $ ^{I+}_BE_{\times} $, $ ^{All}E_{\times} $ can be formed respectively as:
\begin{equation}
\begin{aligned}
& ^{I+}_IE_{\times}=E_{ICP} \times E_{iCLAP} \\
& ^{I+}_BE_{\times}=E_{BoW} \times E_{iCLAP} \\
& ^{All}E_{\times}=E_{ICP} \times E_{BoW} \times E_{iCLAP}
\end{aligned}
\end{equation}

\begin{figure}[htbp]
	\begin{center}
		\includegraphics[width=0.65\columnwidth]{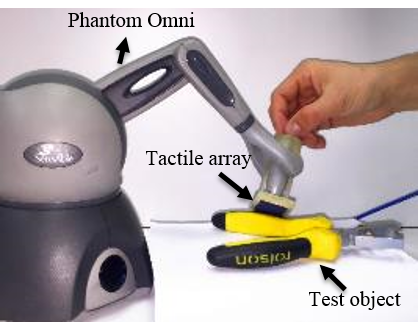}
	\end{center}
	\caption{A Phantom Omni device is used as a
robotic arm to explore the object and a tactile sensor is attached to its stylus.}
	\label{fig:rig.png}
\end{figure}

\section{Data Collection}
As illustrated in Fig.~\ref{fig:rig}, the experimental setup consists of a tactile sensor and a positioning device. A Weiss tactile sensor WTS 0614-34A\footnote{www.weiss-robotics.com/en/produkte/tactile-sensing/wts-en/} was used that has an array of 14$ \times $6 sensing elements. It is of 51mm $ \times $ 24mm and has a spatial resolution of 3.4mm $ \times $ 3.4mm for each element. The sensing array is covered by elastic rubber foam to conduct the exerted pressures. A rate of 5 fps was used as we used in our initial studies \cite{luo2014rotation,luonovel}. A Phantom Omni device\footnote{www.geomagic.com/en/products/phantom-omni/} with six degrees of freedom was used for positioning the tactile sensor and its stylus served as a robotic manipulator. The stylus end-effector position can be obtained that has a nominal position resolution of around 0.055mm. The tactile sensor is attached to the stylus and its surface center is aligned with the end point of the stylus, therefore the end-effector position can be taken as tactile sensor position in the world coordinates.

During the data collection, each object was explored five times. Each exploration procedure was initialized without object-sensor interaction. The stylus was controlled with a speed of around 5 mm/s to explore the object while keeping sensor plane normal to the object surface; in this manner, the object surface was covered while a number of tactile observations and movement data of the tactile sensor were collected. Following \cite{luonovel,schneider2009object}, an uninformed exploration strategy was employed. In total, 8492 tactile images with corresponding contact locations for 20 objects were collected, as shown in Fig.~\ref{fig:objects.pdf}. It can be found that some objects are of similar appearances or spatial distributions. For example, pliers 1 and 2 are of similar size and have a similar frame, whereas they have different local appearances, i.e., the shape of jaws. On the other hand, some objects have similar local appearances but have different spatial distributions, for instance, fixed wrenches 1 and 2.

\begin{figure*}[tbh!]
	\begin{center}
		\includegraphics[width=18cm,height=3.2cm]{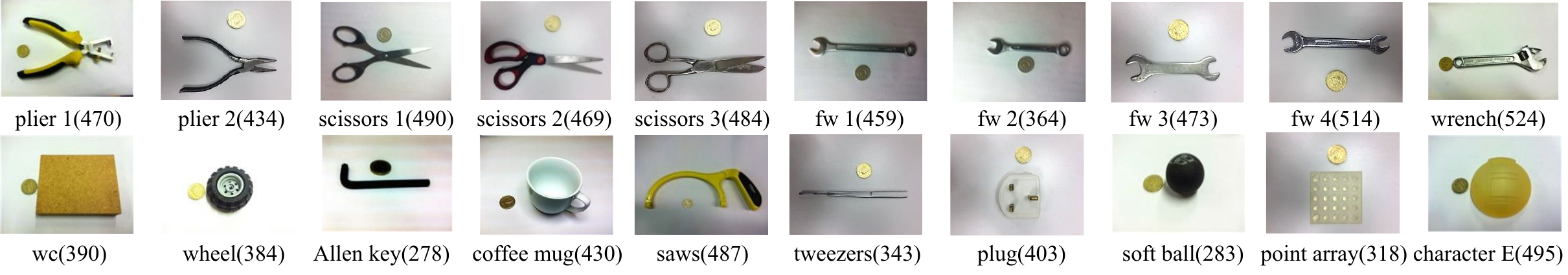}
	\end{center}
	\caption{Objects used for the experiments, labeled from 1 to 20 (left to right, top to bottom). The name and number of collected tactile readings are also given below each object. Note: fw and wc stand for fixed wrench and wooden cuboid respectively.}
	\label{fig:objects}
\end{figure*}

\section{Results and Analysis}
To evaluate the performance of the proposed iCLAP algorithm and extended methods, they are utilized to classify the 20 objects in the experiments. A leave-one-out cross validation is taken and averages of cross validation results are used. The general objective is to achieve a high recognition rate while minimizing the amount of samples needed. Following \cite{luo2016iterative}, the dictionary size \textit{k} is set to 50 through the experiments. In \cite{luo2016iterative}, we also compared different features and Tactile-SIFT features \cite{luonovel} were found to perform best that are therefore also used in this paper.

\subsection{iCLAP vs methods using single sensing modalities}
\begin{figure}[htbp]
	\begin{center}
		\includegraphics[width=1\columnwidth]{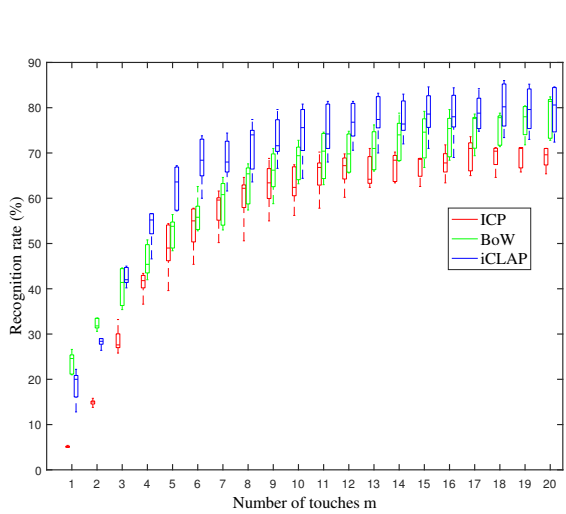}
	\end{center}
	\caption{The recognition rates with ICP, BoW and iCLAP when different number of touches are taken.}
	\label{fig:iclapvssingle}
\end{figure}

The classification results by applying BoW (tactile only), ICP (kinesthetics only) and iCLAP with different number of object-sensor contacts, from 1 to 20, are shown in Fig.~\ref{fig:iclapvssingle}. As the number of contacts increases, all the performances of three approaches are enhanced. When the tactile sensor contacts the test object for less than 3 touches, the tactile sensing can achieve a better performance than the kinesthetic cues as tactile images are more likely to capture key appearance features. In addition, iCLAP outperforms the ICP by up 14.76\%, while performing similarly to BoW. As the number of contacts increases, the performance of iCLAP improves dramatically and it performs much better than those with only one modality. It means that our proposed iCLAP algorithm exploits the benefits of both tactile and kinesthetic sensing channels and achieves a better perception of interacted objects. When the number of contacts is greater than 12, the recognition rates of all the three methods grow slightly and iCLAP still outperforms the other two with single sensing modalities. With 20 touches, an average recognition rate of 80.28\% can be achieved by iCLAP. 

\begin{figure}[htbp]
	\begin{center}
		\includegraphics[width=1\columnwidth]{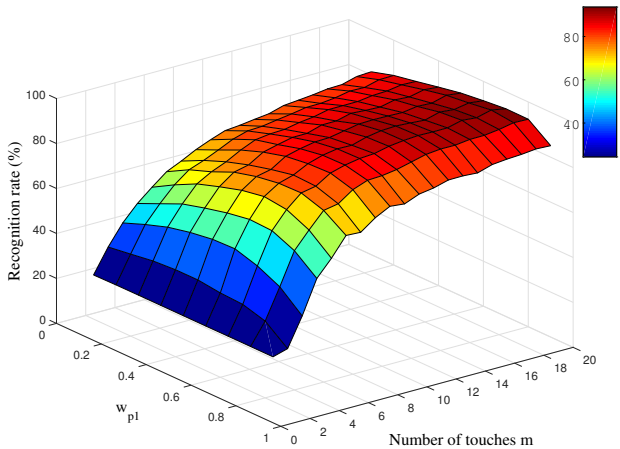}
	\end{center}
	\caption{The recognition rate with different weights $ w_{p1} $ for weighted sum-based decision fusion of ICP and BoW.}
	\label{fig:weight_wp}
\end{figure}
\begin{figure}[htbp]
	\begin{center}
		\includegraphics[width=1\columnwidth]{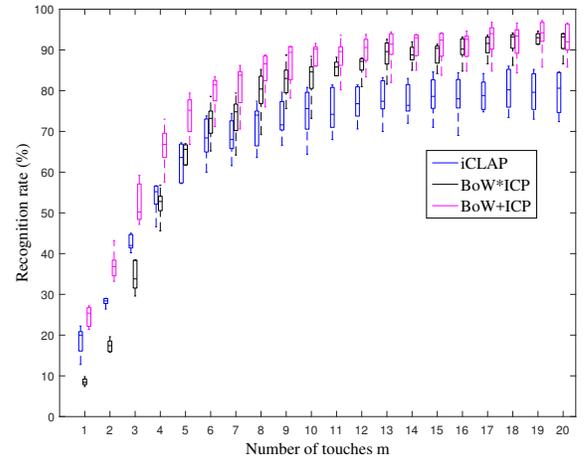}
	\end{center}
	\caption{The recognition rates using iCLAP, product-based (BoW*ICP) and weighted sum-based (BoW+ICP) decision fusion approaches against different number of touches.}
	\label{fig:wp}
\end{figure}

\subsection{iCLAP vs decision fusion approaches}
Both product and weighted sum based decision fusion methods combine recognition decisions of BoW and ICP. The weight $ w_{p1} $ assigned in the weighted sum-based method (Eq.~\ref{eq:decisionfusionweight}) has been investigated to find the optimal combination by brutal force search. The recognition rates with weights $ w_{p1} $ from 0.1 to 0.9 at an interval of 0.1 are shown in Fig.~\ref{fig:weight_wp}. It is found that a good recognition rate of around 90\% can be achieved with 15 touches if $ w_{p1} $ is in the range of from 0.5 to 0.8. And it is observed that the best recognition performance can be achieved with $ w_{p1} $=0.7, i.e.,
\begin{equation}
^I_BE_{+}=0.7E_{ICP}+0.3E_{BoW}.
\end{equation}
As shown in Fig.~\ref{fig:wp}, the weighted sum-based decision fusion method surpasses iCLAP consistently; the product-based decision fusion approach falls behind iCLAP when limited touches are obtained ($<$5) whereas outperforms it when more touches are acquired. Moreover, the weighted sum-based method leads at first ($<$15) yet is caught up by product-based approach when more data are gathered. A good recognition performance, around 90\%, can be achieved by applying either decision fusion approach. It can be estimated that a hybrid fusion strategy, combining decisions of iCLAP and separate sensing sources, can further enhance the classification accuracy.
 
\subsection{Hybrid fusion of iCLAP and classic ICP}
The classification results of iCLAP are first fused with the decisions of classic ICP using only kinesthetic cues to achieve a hybrid conclusion of the object identity. The weight $ w_{p2} $ assigned in the weighted sum-based fusion method (Eq.~\ref{eq:hybridiclapicp}) has been studied using brutal force search, set from 0.1 to 0.9 at an interval of 0.1. The changes of the recognition rate with weight $ w_{p2} $ are shown in Fig.~\ref{fig:weight_pwp.eps}. It is found that a recognition rate of around 80\% can be achieved with 15 touches if $ w_{p2} $ is in the range of from 0.1 to 0.4. And it is observed that the best recognition rate can be achieved with $ w_{p2} $=0.1, i.e.,
\begin{equation}
^{I+}_IE_{+}=0.1E_{ICP}+0.9E_{iCLAP}.
\end{equation}
The recognition rates are shown in Fig.~\ref{fig:pwp.eps}, against the number of contacts. It can be noticed that iCLAP consistently outperforms both the hybrid fusion methods, i.e., product and weighted sum-based approaches. The probable reason is that the inaccurate matching of spatial distributions deteriorates the overall hybrid fusion performance. Nevertheless, the trend of performance enhancement follows that of decision fusion methods, i.e., the weighted sum-based hybrid method outperforms the product-based hybrid approach when limited ($<$5) touches are available whereas the gap is narrowed when more inputs are supplemented.
\begin{figure}[htbp]
	\begin{center}
		\includegraphics[width=1\columnwidth]{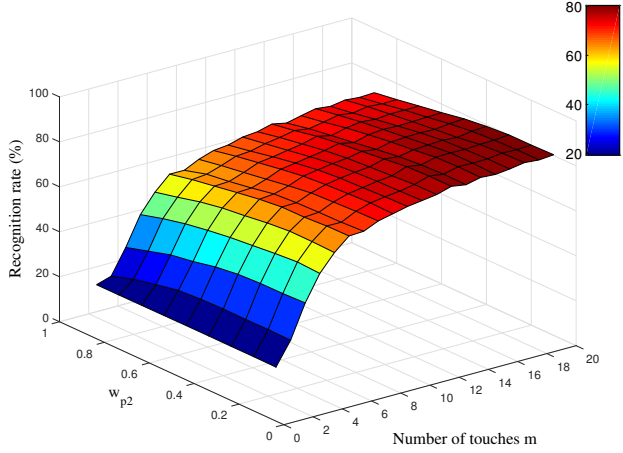}
	\end{center}
	\caption{The recognition rates with different weights $ w_{p2} $ for weighted sum-based decision fusion of iCLAP and ICP.}
	\label{fig:weight_pwp}
\end{figure}

\begin{figure}[htbp]
	\begin{center}
		\includegraphics[width=1\columnwidth]{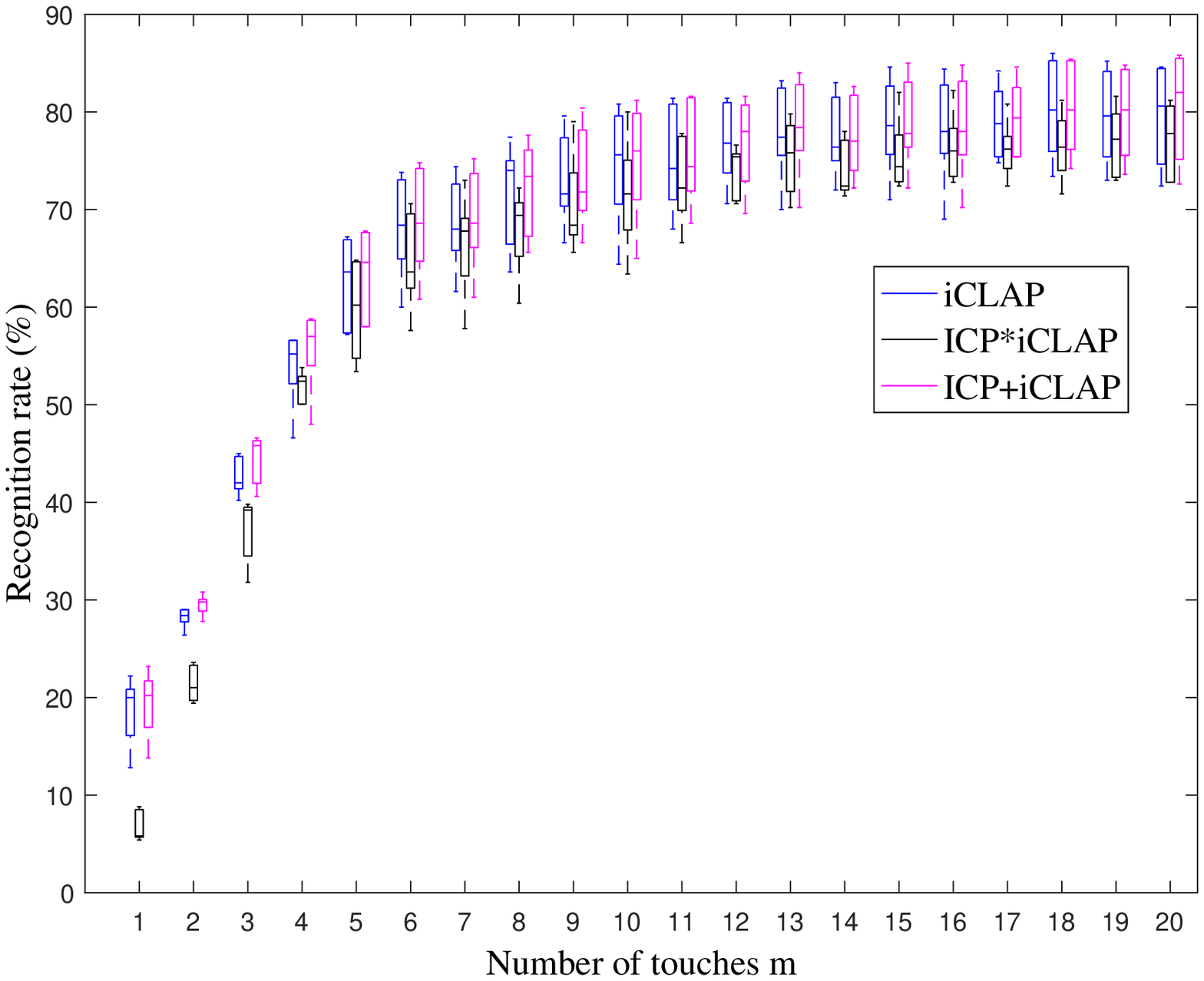}
	\end{center}
	\caption{The recognition rates using iCLAP, product-based (ICP*iCLAP) and weighted sum-based (ICP+iCLAP) hybrid fusion approaches against different number of touches.}
	\label{fig:pwp}
\end{figure} 

\begin{figure}[htbp]
	\begin{center}
		\includegraphics[width=1\columnwidth]{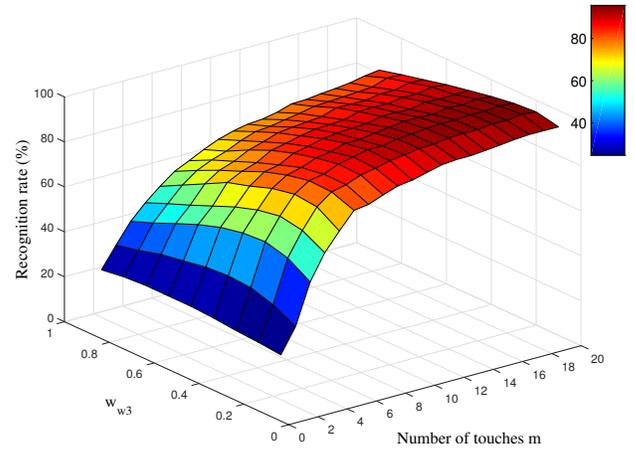}
	\end{center}
	\caption{The recognition rates with different weights $ w_{w3} $ for weighted sum-based hybrid fusion of iCLAP and BoW.}
	\label{fig:weight_wwp}
\end{figure}

\begin{figure}[htbp]
	\begin{center}
		\includegraphics[width=1\columnwidth]{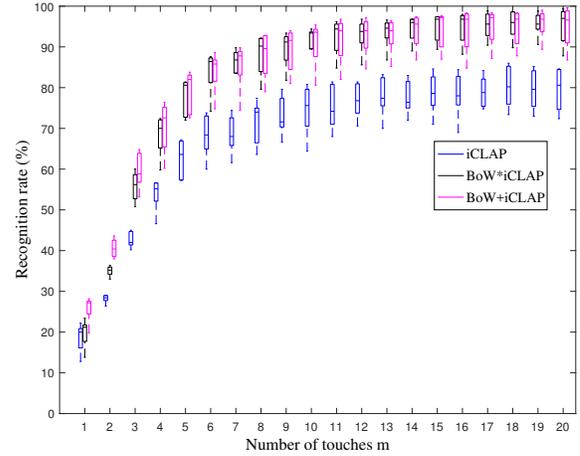}
	\end{center}
	\caption{The recognition rates using iCLAP, product-based (BoW*iCLAP) and weighted sum-based (BoW*iCLAP) hybrid fusion approaches against different number of touches.}
	\label{fig:wwp}
\end{figure}

\subsection{Hybrid fusion of iCLAP and BoW classification pipeline}
The decisions obtained from iCLAP are also combined with those achieved by BoW framework using only local pressure patterns, in a hybrid fusion manner. The weight $ w_{w3} $ assigned in the weighted sum-based method (Eq.~\ref{eq:hybridiclapbow}) has been investigated by brutal force search, set from 0.1 to 0.9 at an interval of 0.1. The changes of the recognition rate with regards to the variance of weight $ w_{w3} $ are illustrated in Fig.~\ref{fig:weight_wwp}. It is found that a good recognition rate of around 90\% can be achieved with 15 touches if $ w_{w3} $ is in the range of from 0.1 to 0.5. And it is observed that the best recognition performance can be achieved when $ w_{w3} $=0.2, i.e.,
\begin{equation}
^{I+}_BE_{+}=0.2E_{BoW}+0.8E_{iCLAP}
\end{equation}
In Fig.~\ref{fig:wwp}, the recognition rates of iCLAP and hybrid methods, both product-based and weighted sum-based, are compared against the number of touches. It can be observed that both hybrid approaches achieve better recognition performance than original iCLAP algorithm consistently, which means the inclusion of the decisions made by BoW framework brings benefit to enhancing the recognition performance of iCLAP algorithm.  Similar to the decision fusion methods, the weighted sum-based hybrid pattern out-stands to a large extent when limited ($<$5) touches are available yet is caught up by product-based hybrid approach when more data are collected. In addition, it can also be found that a satisfactory recognition rate, appropriately 90\%, could be reached with only 10 sensor-object contacts with the hybrid fusion strategy.

\begin{figure}[htbp]
	\begin{center}
		\includegraphics[width=1\columnwidth]{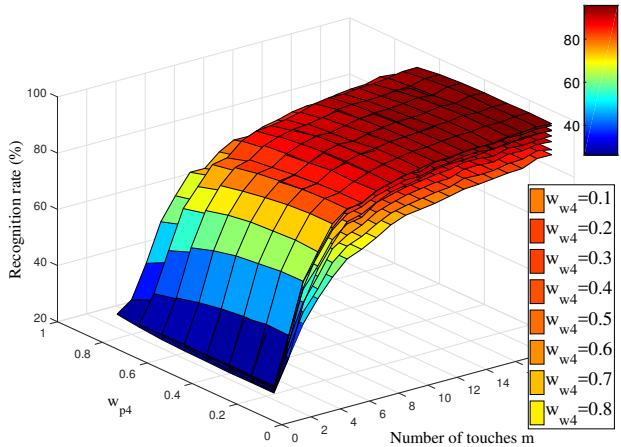}
	\end{center}
	\caption{The recognition rates with different weights $ w_{p4} $ and $ w_{w4} $ for weighted sum-based hybrid fusion of iCLAP, ICP and BoW.}
	\label{fig:weight_wpwp}
\end{figure}
\begin{figure}[htbp]
	\begin{center}
		\includegraphics[width=1\columnwidth]{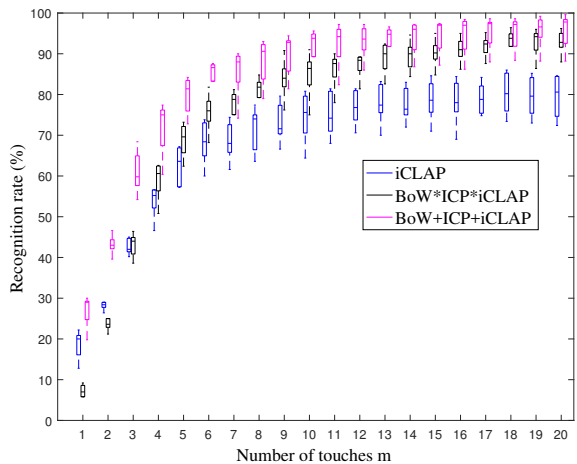}
	\end{center}
	\caption{The recognition rates using iCLAP, product-based (BoW*ICP*iCLAP) and weighted sum-based (BoW+ICP+iCLAP) decision fusion approaches with both tactile and kinesthetic sensing pipelines against different number of touches.}
	\label{fig:wpwp}
\end{figure}

\subsection{Hybrid fusion of iCLAP, ICP and BoW}
The recognition results based on iCLAP are then merged with both decisions acquired by BoW and ICP. As the same, product-based and weighted sum-based hybrid fusion strategies are employed. By employing brutal force search, the weight combinations of $ w_{p4} $ and $ w_{w4} $ assigned in the weighted sum-based method (Eq.~\ref{eq:hybridiclapicpbow}) has been studied, set from 0.1 to 0.8 at an interval of 0.1 with the sum of the weights of three parts as 1. The recognition rates with various weights $ w_{p4} $ and $ w_{w4} $ are shown in Fig.~\ref{fig:weight_wpwp}. It can be seen that the series with $ w_{w4} $=0.2 perform better than ones with the other $ w_{w4} $ values. The combination of $ w_{p4} $=0.2 and $ w_{w4} $=0.2 was found to have the best recognition performance, i.e.,
\begin{equation}
^{All}E_{+}=0.2E_{ICP}+0.2E_{BoW}+0.6E_{iCLAP}
\end{equation}
As shown in Fig.~\ref{fig:wpwp}, the performances of both fusion strategies have been improved compared to the hybrid fusion approaches with only ICP classification pipeline. The probable reason is that the complement of tactile sensing pipeline makes up the inaccurate matching caused by kinesthetic sensing channel. However, the situation alters when compared to the hybrid fusion approaches with decisions of only BoW framework. The recognition rates of weighted sum-based hybrid fusion method are enhanced whereas the performance of product-based hybrid fusion method is deteriorated. The possible reason is that the inclusion of inaccurate kinesthetic based recognition undermines the tactile sensing based classification performance, especially when employing the product-based strategy. A decent recognition rate can be reached with only 10 touches; a comprehensive highest rate of 95.52\% (averaged by cross validation results) can be achieved with 20 sensor-object contacts when a weighted sum based hybrid fusion strategy is employed to integrate decisions of iCLAP, BoW and ICP.

\subsection{Discussions}

\textbf{Multi-dimensional ICP.} The obvious unit mismatch might have a strong effect on the recognition performance using an integer index directly as the 4th dimension of points in the ICP algorithm, e.g., features of clusters 1 and 2 may be more distinct than features of clusters 1 and 7. It is an open issue how to normalize the 4D points due to the different nature of geometrical point coordinates and tactile feature labels. It is proposed that the cluster labels could be ranked by the mutual similarity of different clusters; both the coordinate values and the feature labels could be normalized into the range of from 0 to 1 and weighted during the closest point search. As an alternative way, each data point can be represented as the concatenation of its three positional coordinates with the feature descriptor obtained in this location to form a point in the multi-dimensional space. This method introduces more information of local pressure patterns in the multi-dimensional description of the objects and is potential to improve the algorithm robustness and the recognition performance but this method will introduce more computational burden.

\textbf{Optimal weight assignment in fusion methods.} Based on the experiment results using different fusion methods, it can be found that the weighted sum fusion can achieve a better recognition performance than the product based fusion. However, on the other hand, the complexity of finding an optimal weight assignment is the major drawback of the weighted sum fusion. The brutal force approach is straightforward and easy to be implemented but it brings about additional computational cost and the selection of weights needs to be done by trial and error. The issue of search for appropriate weights for different modalities is still an open question as discussed in \cite{atrey2010multimodal} and beyond the scope of this paper.

\textbf{Haptic exploration.} In the autonomous haptic exploration with the multi-fingered hands with skins, there will have multiple contacts on the object when the object is held in-hand. With the collected tactile patterns and contact locations, the algorithm proposed in this paper can be implemented for the object recognition task. To explore the unknown object, the planning methods in the literature \cite{sommer2016multi,pezzementi2011tactile} can be applied to design the exploration strategies. In the haptic exploration, a confidence level of object recognition can be defined, and multiple grasps are implemented to explore the object surfaces iteratively until the defined confidence level is achieved.

\section{Conclusion and Future Directions}
In this paper we propose a novel algorithm named Iterative Closest Labeled Point to integrate tactile patterns with kinesthetic cues applied in the object recognition task. A bag-of-words framework is first applied to exploit the collected training tactile readings and a dictionary of tactile features is formed by clustering tactile features. By
indexing the dictionary, each tactile feature is assigned a word label. The numerical label is then appended to the 3D location of tactile sensor where it is obtained, to compose a 4D descriptive point cloud of the objects. During the object recognition, the test object point cloud is transformed iteratively to find the optimal reference model in the training set and the distance between them serves as the similarity criterion. The experimental results of classifying 20 objects show that iCLAP can improve the recognition performance (up to 14.76\%) compared to the methods using only one sensing channel. 

We also extend iCLAP to a series of approaches by employing hybrid fusion strategies. The recognition performance has been further improved when the decisions of iCLAP and BoW framework (or also with classic ICP) are integrated using either product based or weighted sum based fusion strategy. Besides, it can be observed that the weighted sum based fusion method outstands when limited number of contacts is available whereas both strategies perform quite well when considerable readings are collected (greater than 10 touches). A satisfactory recognition rate of 95.52\% can be achieved when 20 touches are used and a weighted sum based hybrid fusion strategy is employed to integrate decisions of iCLAP, BoW and classic ICP. 

The proposed iCLAP and its extended approaches can be applied to several other fields such as computer vision related applications. In the view of scene classification \cite{cordts2016cityscapes}, as the landscape observations are correlated with the locations where they are collected, the proposed iCLAP combining the two sensing modalities is expected to enhance the classification performance. It can also be applied to medical applications such as Minimally Invasive Surgeries (MIS). If the local tactile patterns or visual appearances are merged with the spatial distributions of these observations by iCLAP, it is estimated to better recognize the interacted workspace within the body.

There are several directions to extend our work. As only the word label is utilized in the iCLAP to represent the tactile data, there is information loss to certain extent. Therefore, it is planned to include more clues of tactile patterns in the future designed algorithm. In addition, it will also be studied to recognize objects with multiple tactile sensing pads. It is proposed to implement our proposed algorithms on an instrumented robotic hand with multiple tactile array sensors on the fingers and the palm. And it will be explored how to minimize the number of touches to recognize objects in such cases.

\section*{Acknowledgment}
The work presented in this paper was partially supported by the Engineering and Physical Sciences Council (EPSRC) Grant (Ref: EP/N020421/1) and the King's-China Scholarship Council PhD scholarship.

\bibliographystyle{IEEEtran}
\bibliography{references} 

%


\begin{IEEEbiography}[{\includegraphics[width=1in,height=1.25in,clip,keepaspectratio]{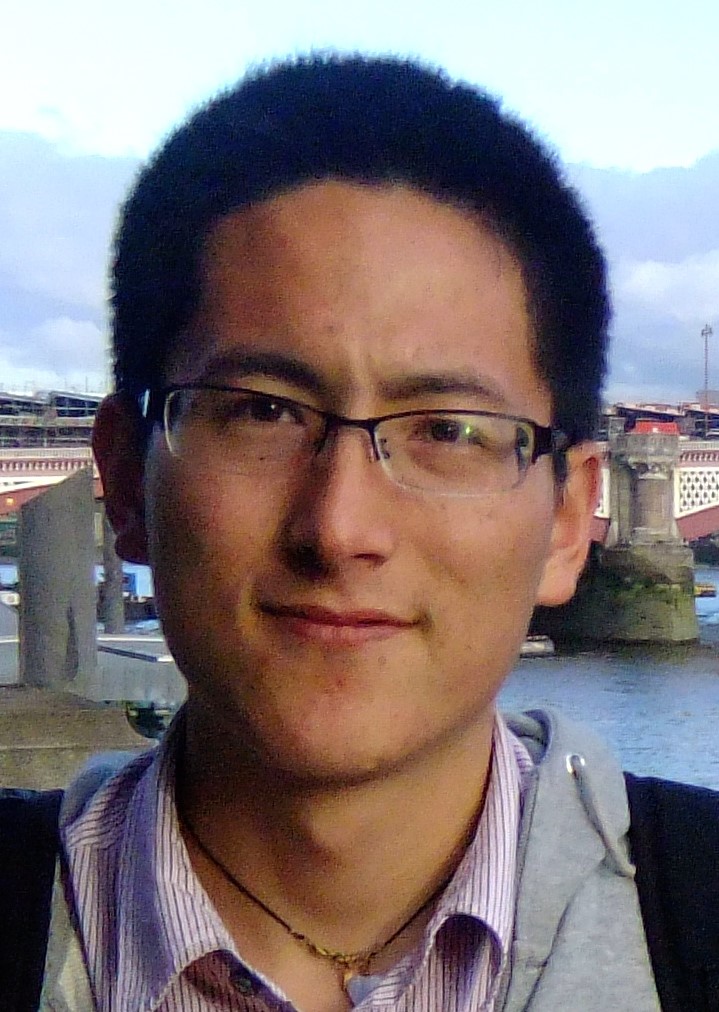}}]{Shan Luo}
is a Lecturer (Assistant Professor) at the Department of Computer Science, University of Liverpool. Previous to Liverpool, he was a Research Fellow at Harvard University and University of Leeds. He was also a Visiting Scientist at the Computer Science and Artificial Intelligence Laboratory (CSAIL), MIT. He received the B.Eng. degree in Automatic Control from China University of Petroleum, Qingdao, China, in 2012. He was awarded the Ph.D. degree in Robotics from King\rq s College London, UK, in 2016. His research interests include tactile sensing, object recognition and computer vision.
\end{IEEEbiography}

 \begin{IEEEbiography}[{\includegraphics[width=1in,height=1.25in,clip,keepaspectratio]{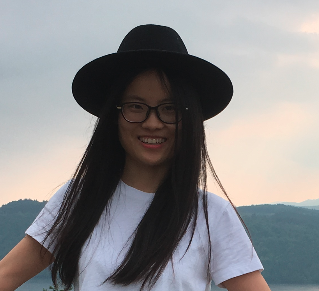}}]{Wenxuan Mou}
 is currently a PhD student in Queen Mary University of London, UK. She received her Master degree in Computer Vision from Queen Mary University of London in 2014 and the B.Eng. degree in Automatic Control from China University of Petroleum, Qingdao, China, in 2012. Her research interests are in the areas of affective computing, computer vision and machine learning.
 \end{IEEEbiography}
 
 \begin{IEEEbiography}[{\includegraphics[width=1in,height=1.25in,clip,keepaspectratio]{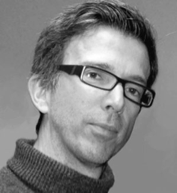}}]{Kaspar Althoefer (M-02)}
is a roboticist with a Dipl.- Ing. degree from the University of Aachen, Germany, and a Ph.D. degree from King\rq s College London, U.K. Currently, he is Professor of Robotics Engineering and Director of ARQ (Advanced Robotics @ Queen Mary) at Queen Mary University of London, U.K., and Visiting Professor in the Centre for Robotics Research (CoRe), King\rq s College London. His research expertise is in soft and stiffness-controllable robots, force and tactile sensing, sensor signal classification and human-robot-interaction, with applications in minimally invasive surgery and manufacturing. He co-/authored more than 250 refereed research papers in mechatronics and robotics.
 \end{IEEEbiography}

 \begin{IEEEbiography}[{\includegraphics[width=1in,height=1.25in,clip,keepaspectratio]{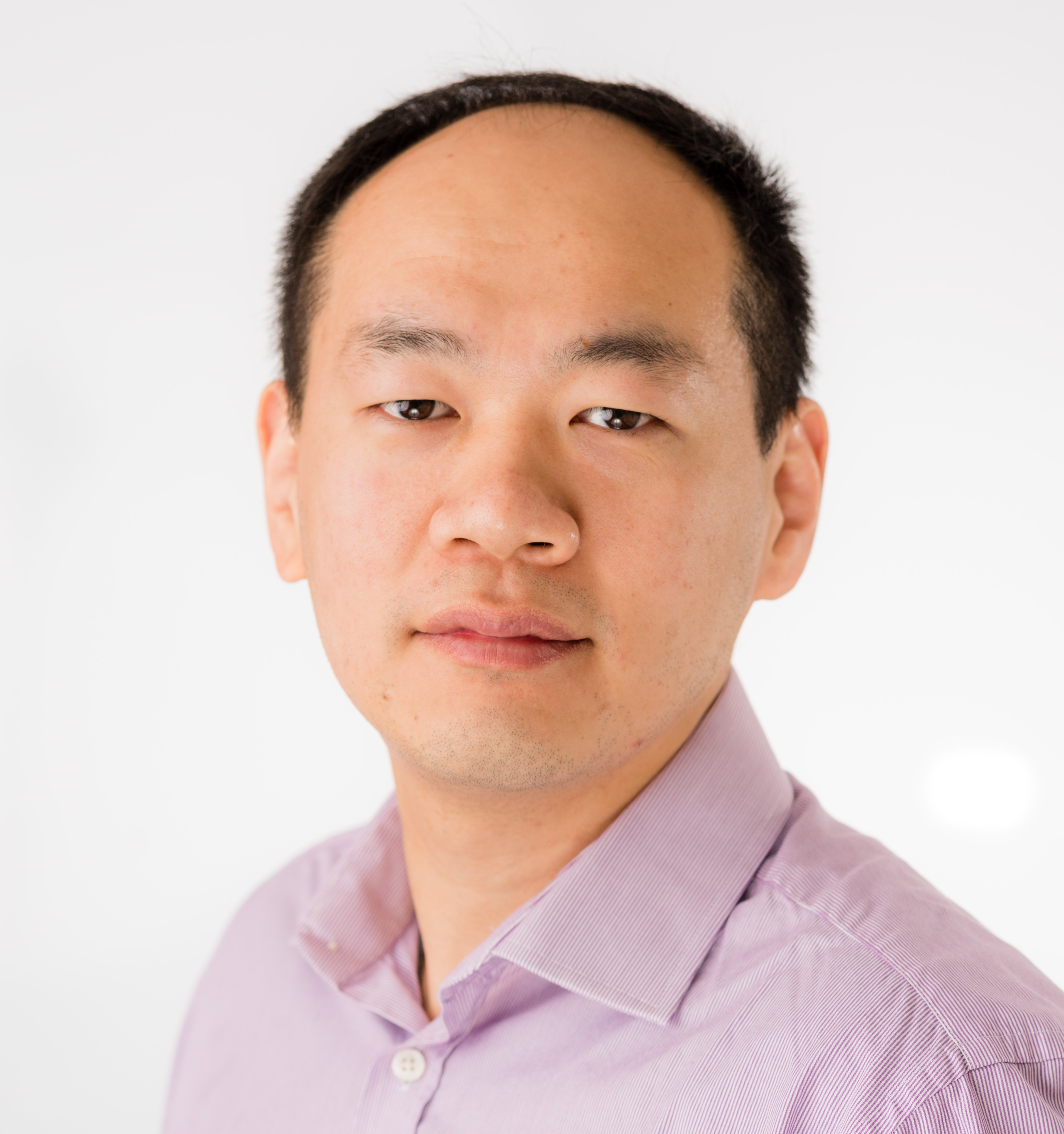}}]{Hongbin Liu (M-07)}
is a Senior Lecturer (Associate Professor) in the Department of Informatics, King\rq s College London (KCL) where he is directing the Haptic Mechatronics and Medical Robotics Laboratory (HaMMeR) within the Centre for Robotics Research (CoRe). Dr. Liu obtained his BEng in 2005 from Northwestern Polytechnical University, China, MSc and PhD in 2006 and 2010 respectively, both from KCL. He is a Technical Committee Member of IEEE EMBS BioRobotics. He has published over 100 peer-reviewed publications at top international robotic journals and conferences. His research lies in creating the artificial haptic perception for robots with soft and compliant structures, and making use of haptic sensation to enable the robot to effectively physically interact with complex and changing environment. His research has been funded by EPSRC, Innovate UK, NHS Trust and EU Commissions. 
 \end{IEEEbiography}



\end{document}